# Spoiled for Choice? Personalized Recommendation for Healthcare Decisions: A Multi-Armed Bandit Approach

**Tongxin Zhou[†], Yingfei Wang[†], Lu Yan[††], Yong Tan[†]**

Michael G. Foster School of Business, University of Washington, Seattle, WA 98195 [†]

Kelley School of Business, Indiana University, Bloomington, IN 47405 [††]

## Abstract

Social media-based platforms tend to provide much information and numerous options for users, making choice overload a prevalent issue in many online communities. Online healthcare communities (OHCs), which provide users with various healthcare interventions to promote healthy behavior and improve adherence, are no exception. When faced with too many choices, however, individuals may find it difficult to decide which behavior intervention to take, especially when they lack the experience or knowledge to evaluate different options. Consequently, the choice overload issue may negatively affect users' engagement in health management in OHCs. In this study, we take a design-science perspective to propose a recommendation framework that helps users to select healthcare interventions. Taking into account that users' health behaviors can be highly dynamic and diverse, we propose a multi-armed bandit (MAB)-driven recommendation framework, which enables us to adaptively learn users' preference variations while promoting recommendation diversity in the meantime. To better adapt an MAB to the healthcare context, we synthesize two innovative model components based on prominent health theories. The first component is a deep-learning-based feature engineering procedure, which is designed to learn crucial recommendation contexts in regard to users' sequential health histories, health-management experiences, preferences, and intrinsic attributes of healthcare interventions. The second component is a diversity constraint, which structurally diversifies recommendations in different dimensions to provide well-rounded support for individuals' health management. We apply our approach to an online weight management context and evaluate it rigorously through a series of experiments. Our results demonstrate that each of the design components is effective and that our recommendation design outperforms a wide range of state-of-the-art recommendation systems. Our study contributes to the emerging IS research on the application of business intelligence. The results have important implications for multiple stakeholders, including online healthcare platforms, policymakers, and users.

# 1  Introduction

Internet technologies enable information to be generated and disseminated at almost no cost, which accelerates the growth of information in online environments. Social media platforms, for example, allow users to share abundant content, including blogs, music, videos, and other formats, which individuals can freely choose to consume. Although various information options increase individuals' choice opportunities, having too many choices can be overwhelming and sometimes even confusing. Often, individuals experience difficulties in spotting content that is truly relevant to themselves or in which they are indeed interested (Konstan and Riedl 2012; Ricci et al. 2015). Such a choice overload issue can lessen users' experience and create barriers to individuals' engagement in online platforms.

Online healthcare communities (OHCs), which are social-media-based platforms that gather users with similar health-management interests, are no exception. Due to their easy access, OHCs are increasingly being used by individuals to learn about their illness, become familiar with treatment routines, and connect with others in similar circumstances. Typical OHCs provide users with various healthcare interventions to promote healthy behavior and improve adherence. Examples include behavioral treatment programs or plans that help individuals to establish healthy habits in regard to diet and physical exercise. During the recent COVID-19 pandemic, for example, individuals often engage in online work-out activities to relieve stress and stay healthy (Pew Research Center 2020). Individuals can freely choose interventions in which to participate in an online environment. When faced with too many choices, however, individuals may find it difficult to decide which option to take, as they may not know what would work or even what to expect, especially when they are not healthcare professionals and do not have adequate experience in evaluating each choice. As a result, they may fall into analysis paralysis (Oulasvirta et al. 2009) and fail to engage in any health-management activities. This may harm their self-intervention adherence and outcome (Nutting et al. 2011; Snyderman and Dinan 2010).

The choice overload issue significantly affects one's participation experience or outcome in OHCs, leading to the pressing demand for services that can better fit individuals' healthcare needs. Therefore, in this study, we aim to follow the design-science paradigm to develop a personalized healthcare recommendation system as a means to support individuals' engagement in health management. Recommendation systems are intelligence-based algorithms that can help users to filter information and discover alternatives that they might not have found otherwise (Konstan and Riedl 2012; Vozalis and Margaritis 2003). Existing recommendation systems deploy various approaches to learn users' preferences from user-behavior data, such as collaborative filtering, content-based filtering, and hybrid models (Zhang et al. 2019). Research has shown that recommendation systems can effectively improve business performance and customer experience (Konstan and Riedl 2012; Pu et al. 2011) in ecommerce settings.

Despite their extensive use in ecommerce settings, whether and how recommendation systems can be integrated with online healthcare platforms has received little attention and remains largely underexplored. There are several unique patterns associated with users' health behaviors that create challenges in healthcare recommendations. First, previous health studies suggest that individuals' health behaviors are frequently affected by their evolving health status and health-management experiences (Johnson et al. 2002; King et al. 2006; Yan and Tan 2014). Thus, individuals' healthcare needs can exhibit strong temporal dynamics. Second, individuals' health management usually contains multi-dimensional effort, as promoting health requires individuals to make a series of changes in all aspects of motivation and lifestyle. For instance, in weight management, individuals need to jointly monitor and manage different behavioral aspects, such as dietary behaviors and participation in physical activities. These patterns indicate that individuals' healthcare needs can be diverse, as individuals may need support for each type of health-management activity.

Given these unique patterns of individuals' health behaviors, conventional recommendation systems that are proven effective in ecommerce settings may not be effective in the healthcare context. This is because these algorithms generally exploit historical data to learn users' preferences. As individuals' health behaviors are continually changing, their health-behavior variations may not be fully captured by the historical data, especially when individuals' health-behavior data remain limited.[1] Thus, a mere exploitation of historical data may not be sufficient in healthcare recommendations. In addition, when individuals do not have well-established preferences about healthcare interventions, they may dynamically form their preferences based on the recommended items. The conventional recommendation systems do not take into account such interactions between users and recommendations and, thus, may not be effective in improving long-term recommendation performance (Liu et al. 2019). Finally, conventional recommendation systems are generally shown to over-specialize recommendations (Fleder and Hosanagar 2009; Pariser 2011). Thus, they may not well support users' diverse healthcare needs.

---

[1] Different from ecommerce recommendation problems, data collection in healthcare recommendations can be a costly procedure, as healthcare is a long-term service, and individuals typically do not generate much health-behavior data when consuming healthcare goods.

These research gaps motivate us to propose a novel recommendation design that utilizes a multi-armed bandit (MAB) as the main building block. An MAB is an online-learning framework in statistics and machine learning for solving decision-making problems in noisy or changing environments (Auer et al. 2002; Chapelle and Li 2011). Specifically, when decision-makers (e.g., service providers) do not know the outcome of an action (e.g., recommendation), an MAB can help them to sequentially select choice alternatives while actively gathering information on each alternative's expected payoff (Zeng et al. 2016). In this process, an MAB strikes a balance between *exploiting* the learned knowledge to gain immediate rewards (reusing a highly rewarding alternative from the past) and *exploring* potential better alternatives (trying new or less-used alternatives to gather more information), which is known as the "exploitation-versus-exploration" tradeoff. By doing so, an MAB aims to maximize the cumulative reward during the entire decision-making period. In the healthcare recommendation context, service providers tend to have little knowledge about users' healthcare preferences, as individuals may constantly change their health behaviors and healthcare needs. Thus, the MAB framework can be used in such a setting to efficiently guide the learning of users' changing healthcare needs. In addition, through the exploration process, an MAB framework can promote the discovery of users' diverse healthcare needs, which may not be revealed by their historical behavior data.

To better adapt an MAB to the healthcare-recommendation context, we follow prominent health-behavior theories to further extend and enhance a standard MAB by synthesizing two model components, deep-learning-based feature engineering and diversity constraint. First, we design and implement two deep-learning models to extract user embeddings and item embeddings, which enables us to capture information that is critical to a healthcare decision-making context, such as users' health histories and health-behavior sequences (Johnson et al. 2002; King et al. 2006; Yan and Tan 2014) and intrinsic attributes of healthcare interventions. Taken together, the constructed user embeddings and item embeddings help to improve the personalization and contextualization of healthcare recommendations. The second model component is incorporated based on social cognitive theory (SCT) (Bandura 1991; Bandura 1998). SCT proposes a classic paradigm for understanding individuals' personal-influenced-based health-management behaviors. Based on SCT, we theorize the major dimensions of health management, and we use a diversity constraint to ensure that recommendations are structurally diversified along each of the health-management dimensions, so that individuals are provided with well-rounded support. To this end, we propose a Thompson sampling (TS)-based algorithm to solve this constrained recommendation task.

Our proposed recommendation framework is evaluated through a series of experiments, using data collected from a leading non-commercial online weight-loss platform in the United States. The focal platform provides weight-loss challenges to users, which are structured behavioral treatment programs to help users to manage short-term weight-loss goals, such as changing a dietary behavior, increasing physical exercise, and reducing weight in certain periods. We apply our recommendation framework to this weight-management setting to help users to find the most relevant challenges as a means to improve their engagement in weight-management activities. Our evaluation results suggest that each of our proposed model components is effective and that our recommendation framework significantly outperforms a wide range of benchmark models, including UCB, $\varepsilon$-greedy, and state-of-the-art conventional recommendation systems, such as context-aware collaborative filtering (CACF), probabilistic matrix factorization (PMF), and content-based filtering (CB). In addition, we demonstrate that our recommendation framework can more effectively learn the dynamics and the diversity distribution in users' challenge choices. From users' perspectives, we find that our recommendation design can serve to benefit a larger user population on the platform. Finally, we take a further step to evaluate our recommendation performance with respect to users' weight-loss outcomes. The evaluation results suggest that our proposed recommendation design can help users to achieve the highest average weight-loss rate compared to the benchmark models.

Our study makes several key contributions to the literature and practice. First, one major contribution of our study is the proposed healthcare recommendation framework, which demonstrates that prescriptive analytics can be integrated via a design-science artifact (Abbasi et al. 2016; Chen et al. 2012) to provide decision-making support for individuals' health management. The novel aspects of our recommendation framework include (1) a deep-learning-based feature engineering procedure, (2) a domain-knowledge-driven diversity constraint, and (3) a customized online-learning scheme. To the best of our knowledge, our study is among the first to combine an MAB with deep context representations and to introduce recommendation constraints for diversity promotion. Second, from a practical perspective, our recommendation framework can be applied to address real-world challenges in healthcare recommendations. Online healthcare platforms can adopt our recommendation design to improve users' health-management experience on the platform. Finally, the design of our recommendation framework can be further generalized to settings beyond healthcare. The online-learning scheme of an MAB enables decision-makers to adaptively adjust their strategies to minimize opportunity cost, and the deep-learning-based feature engineering procedure can help decision-makers to better understand the context-dependency of their decision results.

## 2 Literature Review and Design Theories

Our study is related primarily to two streams of literature, that is, individuals' health management and recommendation systems. In the following, we first review prominent health-behavior theories to identify the unique behavior patterns associated with individuals' health management. This discussion provides the theoretical foundation for our recommendation design. We then review the existing recommendation algorithms and discuss their limitations in delivering healthcare recommendations with respect to individuals' health-behavior patterns. Finally, we introduce an online-learning framework, MAB, which has gathered increasing attention from the literature for its capability of solving decision-making problems under uncertainty. We explain how an MAB framework can be implemented in capturing individuals' health-behavior patterns in the healthcare recommendation process.

### 2.1 Individuals' Management of Health: Dynamics and Diversity

Individuals' lifestyles play a significant role in affecting their quality of health. Poor health behaviors, such as smoking, alcohol abuse, and sedentary living habits, have been shown to be associated with multiple health risks (CDC 2019). Thus, the management of personal health usually requires individuals to invest effort into making a health-behavior change. For example, in managing a chronic condition, such as obesity or type 2 diabetes, patients need to continually self-regulate their ongoing lifestyle in regard to dietary behaviors and participation in physical activities. Researchers have found that patients' active engagement in health management is generally associated with improved adherence to treatment plans and better health outcomes (Nutting et al. 2011; Snyderman and Dinan 2010).

According to prior health-behavior studies and theories (Bandura 1991; Bandura 1998; Johnson et al. 2002), individuals' health management may exhibit unique patterns, such as behavior dynamics and diversity. These patterns play a decisive role in shaping individuals' preferences for healthcare interventions and affect the design of healthcare recommendation systems. In the following, we introduce several prominent health-behavior theories to motivate our recommendation design.

### 2.1.1 Temporal Dynamics of Health Behavior

Previous health studies have generally depicted individuals' health management as a dynamic process. Johnson et al. (2002) suggested that, in the process of health management, individuals may frequently adapt their health behaviors based on their personal health condition and health-management experiences, such as treatment compliance, self-monitoring, and healthcare-knowledge seeking. In addition, the social environment may dynamically transform individuals' health behaviors by affecting mental well-being (King et al. 2006; Yan and Tan 2014). For instance, the exchange of emotional or informational support among peers may encourage optimism and self-esteem of individuals (DiMatteo 2004), which can help them to better comply with a treatment plan and make a behavior change (Johnson and Wardle 2011; Krukowski et al. 2008; Wang et al. 2012). Together, these studies indicate that individuals' health behaviors need to be understood with respect to specific health and social context. As health and social context may evolve with time, individuals' health behaviors generally exhibit strong temporal dynamics.

### 2.1.2 Diversity in Healthcare Needs

Psychosocial theories have extended our understanding of how cognitive and social factors contribute to personal health, among which SCT (Bandura 1991; Bandura 1998) is widely used in the health literature to describe individuals' health-management behaviors. SCT proposes a personal-influence-based self-regulation model, in which individuals exert control over their motivation and behaviors to achieve better health outcomes. The theory suggests that individuals' self-regulation contains multi-dimensional effort. First, individuals need to set proper health goals to motivate themselves toward a desirable health outcome. Second, individuals need to operationalize their goals into actual behavioral aspects so that they can gain behavior-management skills and strategies to tackle challenges and fulfill expectations effectively. Depending on health contexts, individuals may need to attend to different behavioral aspects at the same time. In weight management, for example, individuals need to manage both their dietary behaviors and physical activities to control their calorie intake and expenditure.

Based on SCT, there are two major health-management dimensions: outcome-oriented dimension(s) and behavior-oriented dimension(s). The former influences individuals' motivation for health behaviors, whereas the latter affects the course of behavior execution. Corresponding to these dimensions, individuals may need different types of support to guide them through the self-regulation process. For instance, individuals may need instructions on setting reasonable health goals to help them understand and manage their progress toward a targeted health condition; as well, they may need suggestions on how to cope with difficulties in the process of establishing health-behavior routines. These patterns indicate that individuals' healthcare needs can be diverse.

## 2.2    Review of Existing Recommendation Systems

The dynamic and multifaceted nature of health management has brought new challenges in healthcare recommendations. In this section, we review the existing recommendation systems to discuss the research gaps associated with conventional recommendation schemes that have impeded them from addressing individuals' unique health-behavior patterns.

Recommendation systems are intelligence-based decision-making algorithms that can help users to filter information or product choices based on their own preferences or interests, especially when there is information or product overload (Konstan and Riedl 2012; Vozalis and Margaritis 2003). During the last few decades, recommendation systems have garnered considerable attention from both academia and industry for their capability in delivering personalized services and generating benefits for service providers and customers (Isinkaye et al. 2015; Pathak et al. 2010; Pu et al. 2011). In the literature, a large body of research has focused on batch-learning-based recommendation systems, such as collaborative filtering, content-based filtering, and hybrid models (Zhang et al. 2019). These recommendation systems generally adopt a "first learn, then earn" recommendation scheme. That is, they first learn users' preference patterns based on a series of historical data, and then they fully exploit the learned knowledge to make future recommendations. For example, collaborative filtering makes recommendations based on similarities in users' item-selection histories (Adomavicius and Tuzhilin 2005; Sedhain et al. 2014), and content-based filtering leverages the content attributes of users' previously selected items (Bieliková et al. 2012; Pon et al. 2007). Previous studies have proposed a variety of techniques to learn users' preference patterns from historical data, such as context-aware recommendation systems (CARS) that model contextual dependency of users' behaviors, and model-based techniques to learn latent user representations.

The "first learn, then earn" scheme, however, is based on the assumption that users' preferences have a static pattern that can be well represented by the historical data (Adomavicius and Tuzhilin 2005; Sahoo et al. 2012). When users' preferences are constantly changing, such recommendation methods may become less effective in adapting to individuals' behavior dynamics, as it is likely that individuals' preference patterns will not be fully captured by the data. In addition, prior studies have generally shown that the batch-learning-based models tend to over-specialize recommendations in the long run (Yu et al. 2009), as they tend to focus on well-known items that already have accumulated adequate historical information, whereas the items with limited historical data will be overlooked (Fleder and Hosanagar 2009; Pariser 2011). As a result, these models can be ineffective in satisfying individuals' diverse healthcare interests. These research gaps motivate us to propose an online-learning scheme, i.e., multi-armed bandit (MAB), to address the dynamics and diversity in individuals' health behaviors to improve healthcare recommendations.

## 2.3 Providing Healthcare Support under Uncertainty: An Online-Learning Scheme

In most real-world decision-making scenarios, decision-makers usually do not know the expected utility of an action and can learn only from experience (Cohen et al. 2007; Mehlhorn et al. 2015; Speekenbrink and Konstantinidis 2015). In statistics and machine learning, multi-armed bandit (MAB) has been proposed to explicitly formulate such decision-making scenarios under uncertainty (Auer et al. 2002; Gittins 1979).

Specifically, an MAB models a sequential decision-making problem in which the underlying reward distribution for each action is unknown, and data can be obtained in a sequential order to update knowledge of the reward distribution. The rationale of an MAB algorithm is to adaptively learn the reward associated with each action while gathering as much reward as possible during the entire decision-making process, that is, *earning while learning* (Misra et al. 2019). In order to do so, an MAB strikes a balance between exploration and exploitation (Kim and Lim 2015; Li et al. 2010; Tang et al. 2014). That is, on the one hand, an MAB reuses highly rewarding alternatives from the past to ensure explicit short-term rewards, that is, "exploiting" the environment (Cohen et al. 2007; Mehlhorn et al. 2015); on the other hand, it takes actions to learn the outcome associated with the less-explored alternatives to minimize opportunity cost, that is, "exploring" the environment (Cohen et al. 2007; Speekenbrink and Konstantinidis 2015).

It is worth noting that the online-learning scheme stands in contrast to batch-learning algorithms. The former actively collects data to learn the environment, with a forward-looking goal of maximizing long-term rewards. In other words, online learning may deviate from the current "best" knowledge from time to time in exchange for potential better learning performance and higher rewards collected in the future. In contrast, batch-learning algorithms fully exploit the current knowledge without exploring potential better opportunities that are not shown in the historical data. As such, they tend to interact with the environment in a passive and myopic manner and may not learn effectively when the environment contains many uncertainties that cannot be represented by current data.

Research has shown that online-learning algorithms, i.e., MABs, are suitable for tackling decision-making problems in noisy and changing environments (Speekenbrink and Konstantinidis 2015). For example, Misra et al. (2019) applied an MAB to a pricing problem in which the volume of demand was uncertain. Schwartz et al. (2017) used an MAB to improve advertising design when online advertisers were not able to identify targeted users. In our healthcare-recommendation context, service providers (e.g., online healthcare platforms) usually have little knowledge of users' healthcare needs or preferences, especially when users frequently change their behavior patterns. An MAB can be used in such a setting to help service providers to effectively explore users' preference variations while improving users' online engagement during the process. In addition, through exploration, an MAB increases choice stochasticity and, thus, can better promote recommendation diversity (Qin et al. 2014). Despite these advantages, MABs are seldom studied in healthcare recommendation problems. In this study, we enrich the healthcare recommendation literature by designing an MAB-driven framework for providing personalized healthcare interventions.

## 3 Healthcare Recommendation: Deep Learning and Multi-Armed Bandit

We propose a deep-learning and diversity-enhanced MAB framework for recommending healthcare interventions to address the challenges and research gaps presented in the previous section. First, we adopt an MAB as the main building block of our framework, as it can effectively explore variations in users' healthcare preferences and promote recommendation diversity at the same time. To better adapt an MAB to the healthcare recommendation setting, we then further enhance our framework by synthesizing two model components, that is, deep-learning-based feature engineering and a diversity constraint. As suggested by prior health studies (Johnson et al. 2002; King et al. 2006; Yan and Tan 2014), individuals' health behaviors are dynamically affected by a series of contexts, including their evolving health status, health-management experiences, and social context. Based on these studies, the sequential information embedded in individuals' health histories and health-behavior paths can play an essential role in shaping individuals' health behaviors. Deep-learning models can effectively capture patterns from dynamic temporal sequences and extract complex synergies between different features, thereby enabling the enhanced representation of variations in individuals' health behaviors. We thus incorporate a deep-learning-based feature engineering procedure to improve recommendation personalization and contextualization. In addition, SCT suggests that individuals' health management may contain multi-dimensional efforts. The diversity constraint helps us to structurally diversify recommendations along each theory-driven health-management dimension so that individuals are provided with well-rounded support.

In Figure 1, we provide a graphical illustration of our recommendation design. Each construct of the recommendation framework is intended to enhance healthcare recommendation performance. In a recommendation cycle, we first use deep-learning models to construct representations for users and items, i.e., the user embeddings and item embeddings. We then use the constructed embeddings to capture the contextual features of the recommendation environment, which enables us to learn the context-dependency of the recommendation results and generalize users' feedback. The MAB algorithm, shown on the right side of Figure 1, adaptively learns users' preferences by balancing the exploitation-versus-exploration tradeoff. The diversity constraint seeks to diversify the recommendations along the theorized health-management dimensions. We elaborate each of these constructs in the remainder of this section.

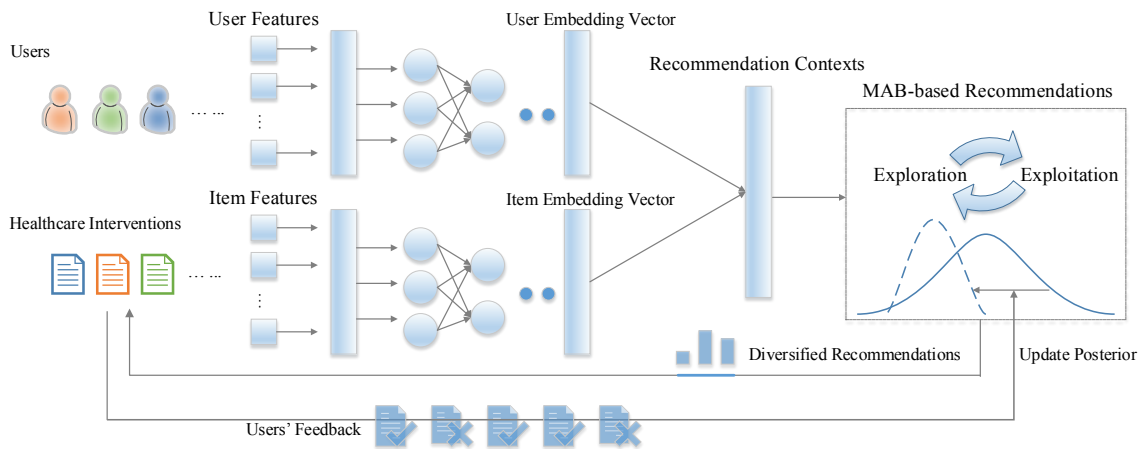

**Figure 1** An MAB-Driven Framework for Healthcare Recommendation

## 3.1 Optimizing Intervention with Multi-Armed Bandits

OHCs provide healthcare interventions to encourage and instruct individuals' health behaviors. In Table 1, we provide several examples of typical healthcare interventions provided in OHCs. We consider the setting in which an online healthcare platform provides intervention suggestions to users on a regular (e.g., weekly) basis. It is unlikely that every individual user will prefer the same interventions, and there is interpersonal heterogeneity in terms of which interventions to adopt and to what extent. The goal of the platform is to adaptively suggest $K$ interventions with the highest chance of improving individuals' engagement in their healthcare management.

| **Table 1** Example of Healthcare Interventions |
| --- |
| 1.   30 minutes jogging every day |
| 2.   No food after 7 p.m. |
| 3.   Avoid chips, soda and alcohol for 2 months |

Formally, let $T$ be the number of recommendation periods and $I$ be the number of users on the platform. Suppose that each time the platform provides each user with $K$ alternatives from a full available item set $C_t$. Let $S_{it}$ denote the subset of items provided to user $i$ at period $t$, $S_{it} \subseteq C_t$. $C_t$ may vary with time. At the end of each period, the platform receives users' feedback on the recommendations, that is, whether they have adopted or engaged in the recommended interventions. Let $r_t(i,k)$ denote user $i$'s feedback on item $k$. Users' feedback serves as a reward for the platform's recommendation decisions; the platform may use the information of users' feedback to update its knowledge about users' preferences and adjust its subsequent recommendations.

We formulate the above recommendation problem as a contextual MAB. In a contextual algorithm, the decision of choice is leveraged upon a set of contextual features of the environment, such as the attributes of choice alternatives and user characteristics, so that the algorithm can exploit the similarity between choice alternatives and deliver online personalized recommendations (Zeng et al. 2016). Contextual MABs learn to map the contexts into appropriate actions (Greenewald et al. 2017). Thus, they are able to personalize recommendations based on specific situations. In addition, as the pool of users and healthcare interventions may likely undergo frequent changes, it is desirable to learn a feature-based model that can generalize users' behavior histories to the user-item pairs that have never or rarely occurred in the past. To this end, in the healthcare recommendation context, we consider two sets of contextual information:

individuals' health-management contexts $\mathbf{x}_{it}$ and attributes of healthcare interventions $\mathbf{z}_k$. We assume that users' feedback, i.e., the intervention engagement decision $r_t(i,k)$, is stochastically generated by an underlying probability that depends on the contexts. We model this probability as a logistic function:

$$E[r_t(i,k)\,|\,\mathbf{v}_{itk}] = (1 + \exp(-\boldsymbol{\theta}_*^T \mathbf{v}_{itk}))^{-1}, \tag{1}$$

where $\mathbf{v}_{itk} = \{\mathbf{x}_{it}, \mathbf{z}_k\}$, and $\boldsymbol{\theta}_*$ denotes the underlying coefficient vector, which can be learned adaptively in the recommendation process. The objective for the online healthcare platform is to maximize the expected cumulative user engagement during the entire course of recommendation, i.e.,

$$\max_{S_{it}} \sum_{t \in [T]} \sum_{i \in [I]} \sum_{k \in S_{it}} E[r_t(i,k)\,|\,\mathbf{v}_{itk}]. \tag{2}$$

## 3.2   Diversity Constraint

Modern recommendation systems should be well-diversified, motivated by the principle that recommending redundant items leads to diminishing returns on utility. In the context of healthcare recommendations, the major health-management dimensions that we identify based on SCT include the outcome-oriented dimension(s) and behavior-oriented dimension(s). Whereas the former helps individuals to gain outcome-driven motivation, the latter enables them to acquire health-management skills and strategies. Thus, to provide individuals with well-rounded support, recommendations need to cover each of the health-management dimensions.

Although an MAB framework is able to promote recommendation diversity through the exploration process, we further incorporate a diversity-constraint MAB to ensure that the exploration is conducted in guided directions and that the recommendations are structurally diversified along each of the health-management dimensions. Formally, our diversity constraint can be expressed as

$$D = \left\{ \left| S \cap dim_{outcome1} \right| \geq 1, \left| S \cap dim_{outcome2} \right| \geq 1,\quad, \left| S \cap dim_{behavior1} \right| \geq 1, \left| S \cap dim_{behavior2} \right| \geq 1,... \right\}, \tag{3}$$

where $S$ denotes the recommendation set, $dim_{outcome}$ denotes the outcome-oriented dimension(s), and $dim_{behavior}$ denotes the behavior-oriented dimension(s). We subject the optimization problem in (2) to the diversity constraint $D$ to ensure that the recommendation set $S_{it}$ contains suggestions for each health-management dimension.

To solve this constrained recommendation task, we propose an algorithm that is adapted from Thompson Sampling (TS). TS is a machine-learning algorithm that addresses the exploitation-versus-exploration tradeoff presented in a bandit problem. TS is best understood in a Bayesian setting in which it computes the posterior distribution of the unknown parameters $\boldsymbol{\theta}$ in the likelihood function, given the realized stochastic feedback. The rationale of TS is to encourage exploration through probability matching. That is, in each round, a TS algorithm randomly draws alternatives according to its probability of being optimal. Research has shown that TS generally has better empirical performance than do alternative bandit algorithms, such as UCB and $\varepsilon$-greedy (Chapelle and Li 2011). Our algorithm extends an ordinary TS by integrating a constrained optimization problem to solve for the optimal recommendation decisions subject to the diversity constraint. We present the details of our algorithm below.

---

**A TS Algorithm with Diversity Constraint**

**Input**: prior mean $m_j$ and prior variance $\upsilon_j$ for each parameter $\theta_j$, $j = 1, 2, \quad d$.

**For** $t = 1, 2, ..., T$ **do**

    **For** $i = 1, 2, ..., I$ **do**

      *Step 1 (Random Draw)*:

      Draw $\hat{\theta}_j$ from $N(m_j, \upsilon_j)$. $\hat{\boldsymbol{\theta}} = (\hat{\theta}_1, \hat{\theta}_2, ..., \hat{\theta}_d)$.

      **For** arm $k \in C_t$ **do**

        Compute expected reward: $\hat{r}_t(i, k) = (1 + \exp(-\mathbf{v}_{itk}^T \hat{\boldsymbol{\theta}}))^{-1}$.

      **End for**

      *Step 2 (Optimization)*:

      Solve the following optimization problem:

$$\max_{q_k, k \in C_t} \sum_{k=1}^{K} \hat{r}_t(i, k) q_k$$

$$s.t. \quad \sum_{k \in dim_{outcome}} q_k \geq 1, \quad \forall i, t \qquad \text{Diversity constraints}$$

$$\sum_{k \in dim_{behavior}} q_k \geq 1, \quad \forall i, t$$

$$\sum_{k \in C_t} q_k \leq K \qquad \text{Recommendation size constraint}$$

$$q_k = 0 \ or \ 1, \ \forall k \qquad \text{Binary decision}$$

      Offer item $k$ to individual $i$ if $q_k^* = 1$. $S_{it} = \{\forall k, \ q_k^* = 1\}$.

    **End for**

    Observe a new batch of data $(\mathbf{v}_{itk}, r_t(i, k))$, $i \in [I]$, $k \in S_{it}$

    Update the posterior mean by:

$$\mathbf{m} = \arg\min_{\boldsymbol{\theta}} \frac{1}{2} \sum_{j=1}^{d} \upsilon_j^{-1} (\theta_j - m_j)^2 + \sum_{i=1}^{I} \sum_{k \in S_{it}} \log(1 + \exp(-r_t(i, k) \boldsymbol{\theta}^T \mathbf{v}_{itk})).$$

    Update the posterior variance by:

$$\upsilon_j^{-1} = \upsilon_j^{-1} + \sum_{i=1}^{I} \sum_{k \in S_{it}} v_{jikt}^2 p_{ikt} (1 - p_{ikt}),$$

    where $p_{ikt} = (1 + \exp(-\mathbf{m}^T \mathbf{v}_{ikt}))^{-1}$, $v_{jikt}$ is the j-th element of $\mathbf{v}_{ikt}$.

**End for**

---

### 3.3    Deep-Learning-Based Feature Engineering

To improve the characterization of individuals' health-management contexts and enhance recommendation personalization, we design a deep-learning model to construct user embeddings. Specifically, our user-embedding model leverages information on users' attribute features (e.g., gender, age, etc.), health-status trajectories, and health-management behavioral sequences. For each user, the attribute variables usually remain unchanged over time, whereas health status and behavioral sequences will vary with time. Hence, the user embeddings depend on both user $i$ and time $t$ to reflect the evolving dynamics. To properly guide the learning on these aspects, we propose a novel wide-and-deep neural network. The wide-and-deep structure was originally proposed for user response modeling in mobile apps. It combines two branches of user features (i.e., a "wide" branch and a "deep" branch) to facilitate user representation learning (Cheng et al. 2016). In this study, we design a "wide" branch to process users' attribute features to take into account that certain intervention suggestions can be more actionable for specific users given their personal attributes, and we apply a fully-connected structure to account for possible interactions among the attribute features.

We then use a "deep" branch to learn sequence features, such as users' health-status trajectories, historical healthcare-intervention adoptions, and other health-management experiences in regard to self-monitoring activities and social behaviors. The sequence of historical intervention adoptions is included to capture the dynamics in users' preferences. Together with the health-status trajectories, it captures users' evolving health histories and the corresponding changing healthcare preference. In addition, based on prior health theories mentioned in Section 2.1, health-management experiences, such as social supports and self-monitoring activities, may also affect individuals' health behaviors and thus influence their preferences for healthcare interventions. Therefore, we further include related behavior paths to capture the effect of health-management experiences on users' intervention-adoption behaviors. We use Long Short-Term Memory (LSTM) with a self-attention mechanism to capture the dynamically changed patterns in these features and their correlation with adopted healthcare interventions. To address the fact that different sequence features have different dimension scales (e.g., number of social activities vs. numerical intervention attributes), we propose a self-organizing LSTM module and add a balancer in the LSTM cell to tackle unbalanced weights between different input features. The output of the deep model is the embeddings of the adopted healthcare interventions, with the loss function defined as the cosine distance between the last hidden layer (user embedding) and item embedding.

In addition, we enhance the learning process of the deep branch by an auxiliary loss function with the healthcare outcome as the goal. We use the auxiliary loss function to incorporate the effects of healthcare outcomes on individuals' health behaviors and individuals' preferences on healthcare interventions. This is unique to the healthcare recommendation context, in which individuals' health behaviors are fundamentally driven by the goal of optimizing health outcomes. Meanwhile, the auxiliary loss branch can also guide the neural network to properly extract signals from the sequence features, as, otherwise, the gradient flow will not be balanced and the shallow structure will dominate the gradient flow. An illustration of the proposed deep learning architecture for user embedding construction is provided in Figure 2.

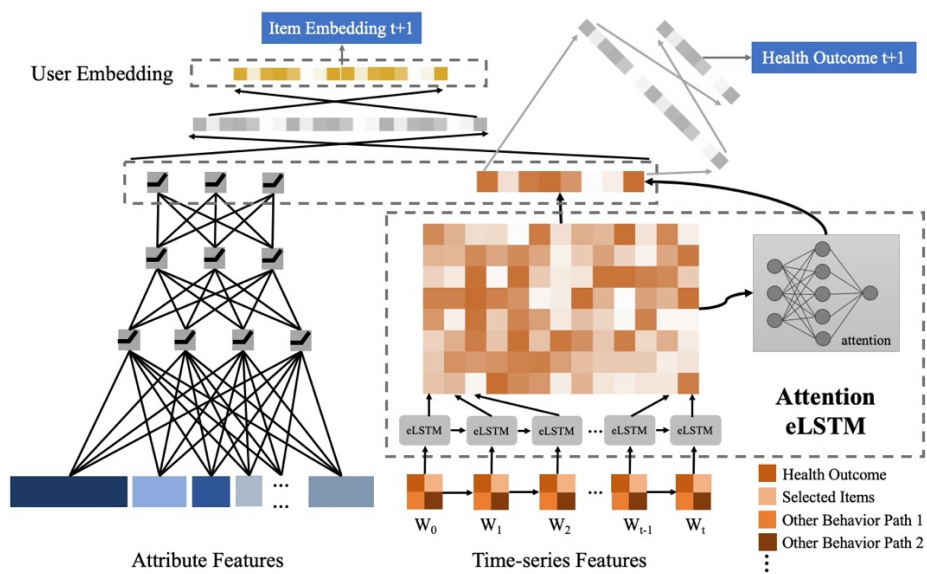

**Figure 2** Illustration of the Base Architecture

Finally, as healthcare interventions provided online are usually in the form of textual information, we build another deep-learning model to construct item embeddings to extract the inherent semantic features embedded in intervention information. For example, a typical online healthcare intervention may contain a short title that highlights the main features of the intervention and a description or instruction that describes the detailed execution procedure. To capture the semantics embedded in this information, we propose a hybrid model, in which we apply LSTM to learn the semantics of the intervention descriptions, and we use average token-level embedding to extract signals from the intervention titles. The outputs are the meta attributes of the intervention (e.g., duration, category, and/or intensity of the intervention). We further fine-tune the token-level embedding in the procedure of representation learning to ensure low information loss.

In sum, the user embeddings and item embeddings help us to learn key contextual information for healthcare recommendations concerning users' health behavior contexts and item attributes, which are then used as the input of our bandit recommendation model. Due to the space limit, more details on our user-embedding and item-embedding models are provided in Appendix A1.

# 4 Evaluation Design

## 4.1 Focal Evaluation Context

To evaluate the performance of our recommendation framework, we collected data from a leading non-commercial online weight-loss platform in the United States. Weight management has garnered considerable attention from healthcare communities due to the increasing prevalence of overweight and obesity in the global populations. Research indicates that overweight and obesity are associated with multiple health risks, such as type 2 diabetes, cardiovascular diseases, and cancer (WHO 2018). Recently, the CDC (2020) announced that obesity is a risk factor for severe COVID-19 complications. Although there are genetic and metabolic influences on body weight, overweight and obesity are likely to occur when individuals develop unhealthy behaviors, such as poor dietary habits and a lack of physical activities, which impede individuals from balancing their calorie intake and expenditure (Franz 2001; Hall and Kahan 2018).

To help users to establish a healthy living style, the focal platform provides weight-loss challenges, which are behavioral treatment programs that help users to focus on a specific weight-loss goal in a short time period. Examples of weight-loss challenges include diet-oriented challenges, such as "cut off processed carbs and include 100g of mixed veg in every meal," and activity-oriented challenges, such as "30 minutes of jogging every day." The diet-oriented and activity-oriented challenges provide behavioral guidance for individuals' weight-management routine. Users can also find weight-loss-oriented challenges that help them to set goals directly for weight changes, such as losing a certain amount of weight during specific periods. Participation in weight-loss-oriented challenges can help users to establish outcome-driven motivation. Each weight-loss challenge is defined by a short title and description that contains information on the challenge goal, duration, and instructions. Users can choose to join any challenge as long as its starting date has not passed. In Appendix A2, we provide a screenshot of the challenge webpage to show how users can retrieve challenge information from the online platform.

The focal platform does not incorporate any recommendation system to facilitate users' challenge selection. During our investigation period, there were more than 100 challenges provided to users. Users may likely find it difficult to select challenges to join, as users may lack the ability to discern, from various choice alternatives, what challenges are suitable for their weight management. In addition, a significant search cost is expended, as users need to spend time reading the challenge descriptions and/or instructions before deciding which challenge to join. These problems can potentially be solved by providing personalized challenge recommendations to support healthy behaviors. Our investigation on weight-loss challenge recommendations can help to improve the match between individuals and weight-loss challenges and, thus, improve individuals' weight-management performance. From the platform's perspective, the recommendations can enhance users' participation experience and, thus, contribute to user maintenance and platform sustainability.

## 4.2 Data

We collected three datasets to support our investigation of recommendation performance. The first dataset contains descriptive information for each challenge provided on the platform. During our data collection window,[2] there were 165 challenges provided on the platform in total. For each challenge, we collected the title, description, and duration. On average, users can choose from about 50 challenges each week. The second dataset is users' challenge-selection histories; that is, we recorded for each user the challenge(s) that he or she selected per week. This dataset enables us to learn users' preferences for weight-loss challenges.

The third dataset contains auxiliary information for each user, such as gender, age, membership duration, initial weight when first joining the platform, online weigh-in activities, the number of friends, and the posts published in the community forum. We use this information to capture users' heterogeneous weight-management contexts. In particular, gender and age are two factors that directly affect individuals' weight status. Membership duration measures users' overall weight-management experiences on the platform. Initial weight and weekly weigh-in records help us measure individuals' weight-loss status and health histories. The number of friends and forum posts provide proxies for the amount of social support available to individuals (Shumaker and Brownell 1984; Yan 2018); thus, we use them to capture the social contexts of users' weight management. We provide a summary of key data statistics in Appendix A3.

---

[2] Our data collection period was January 1, 2014 to April 30, 2014.

### 4.2.1 Users' Challenge-Selection Behaviors

Users on the focal platform, on average, chose two challenges per week. When users chose any challenge, they chose multiple challenges about 70% of the time. As noted, there are three major challenge types on the platform: weight-loss oriented, diet oriented, and exercise oriented. We find that users tend to choose different types of challenges whenever they choose multiple challenges. Specifically, users choose more than two challenge types 92% of the time when they choose multiple challenges; they choose all three types of challenges about 51% of the time. These results indicate the existence of diversity in users' preferences for weight-loss challenges. In addition, we find that users' selection of challenge types drifts over time. That is, users may have preferred certain challenge types at the beginning of a time period and gradually shift to other challenge types as time goes by. These findings provide evidence for the dynamics of users' preferences, which may be due to users' transitions to different weight-loss statuses, in which they need different types of support. It is also likely that users gradually establish their personal tastes in regard to weight-loss challenges during the process of challenge participation. These findings thus provide support for our recommendation design.

### 4.2.2 Challenge Meta Attributes

Weight-loss challenges are presented in a textual format with a title and a description. They aim to help users to manage short-term weight-loss goals, such as changing a dietary behavior, increasing physical exercise, and reducing weight. Goal setting can reinforce individuals' motivation, and well-structured goal formulation can have positive and directional effects on individuals' task performance (Les MacLeod EdD 2012; Locke and Latham 1990). The SMART metric (i.e., specific, measurable, attainable, relevant, and time-bound) has been widely used as a gold standard in areas such as education and healthcare for assessing the quality of goals (Doran 1981; Ogbeiwi 2018). This metric can help individuals to clearly identify the direction for logical action planning and implementation (Ogbeiwi 2017; Ogbeiwi 2018). Thus, SMART-related goal characteristics can influence how individuals perceive the effectiveness of a goal and affect their choice-making behaviors in deciding which goal to pursue.

We use the SMART metric to characterize each challenge based on the challenge description data. As the number of challenges is large and users' challenge-selection data are comparatively sparse, we need to quantify the similarities among challenges, and the SMART-based features can properly guide our calibration of challenge similarity. In particular, corresponding to the goal-setting dimensions specified by the SMART metric, we construct the following meta attributes for each challenge: whether the challenge is specifically defined (specificity), whether the challenge goal is measurable (measurability), the intensity level of the challenge (attainability), whether the challenge is related to diet or physical activity (relevancy), and the time span of the challenge (duration). In Table 2, we provide a summary of the annotated challenge meta attributes, which will be used for learning challenge-embedding representation and downstream recommendation task.

**Table 2** Challenge Description Annotation

| Features | Definition |
|---|---|
| *Specific* | Whether a challenge is specifically defined (0 or 1) |
| *Measurable* | Whether a challenge goal is measurable (0 or 1) |
| *Diet* | Whether a challenge is related to dietary behaviors (0 or 1) |
| *Intensity_Diet* | Intensity level for a diet-oriented challenge (L, M, H) |
| *Activity* | Whether a challenge is related to physical activities (0 or 1) |
| *Intensity_Activity* | Intensity level for an activity-oriented challenge (L, M, H) |
| *Weight-Loss* | Whether a challenge contains a goal for weight changes (0 or 1) |
| *Intensity_Weight_Loss* | Intensity level for a weight-loss-oriented challenge (L, M, H) |
| *Motivational* | Whether a challenge contains motivational words/sentences (0 or 1) |
| *Self-Monitoring* | Whether a challenge requires individuals to regularly monitor and report their weight-loss progress, e.g., body weight, daily diet, running mileage (0 or 1) |
| *Duration* | Time span (in weeks) of a challenge |

In addition to SMART-based attributes, we consider two other features that may affect individuals' engagement in challenge participation: *Motivational* and *Self-Monitoring*. *Motivational* characterizes challenges from the perspective of goal statement, which has been suggested to be important in helping individuals to build up inner motivation (Locke and Latham 1990). *Self-monitoring* is an important step in goal fulfillment, as it helps individuals to process their performance toward goal achievement (Bandura 1991). We use *Self-Monitoring* to indicate whether a challenge encourages individuals to regularly monitor and report their weight-loss progress. The detailed annotation procedure is provided in Appendix A4.

## 4.3   Operationalizing Model Components: Embeddings and Diversity Constraint

### 4.3.1   Operationalizing User and Challenge Embeddings

As noted in Section 3.3, our user-embedding model adopts a wide-and-deep network structure, in which we use the wide branch to capture users' attribute features and the deep branch to capture the sequence features. In our evaluation context, users' attribute features include gender, age, initial weight, and membership duration. The sequence features include three parts. The first part captures users' health status, that is, their historical weight variations. The second part is the sequence of historical challenges chosen by individuals, which we use to account for users' personal tastes. The third part concerns users' other behavioral sequences, such as their past social activities (e.g., establish friendships with other users and publish forum posts) and self-monitoring activities (e.g., weigh-in). The auxiliary loss head is designed to measure the weight loss in the next time period, where we choose a combined loss of MSE for absolute value prediction and cross-entropy loss for weight-loss sign prediction. We present the detailed network structure and the loss functions in Appendix A1.

For our challenge-embedding model, we use challenge name and description as the inputs, and the annotated challenge meta attributes based on the SMART metric are the outputs. The annotated challenge meta attributes help us to depict the key characteristics of a weight-loss-related goal; thus, they provide a good standard for calibrating challenge similarity in our focal context.

### 4.3.2 Operationalizing Diversity Constraint

In operationalizing our diversity constraint, we identify weight loss as the outcome-oriented dimension, as it is the health goal that individuals aim to achieve in our focal context. We identify diet and physical exercise as two behavior-oriented dimensions, as they are two essential behavioral-regulation aspects in weight management. With respect to these dimensions, we ensure that our recommendations cover all three challenge types, i.e., weight-loss oriented, diet oriented, and exercise oriented. Therefore, our diversity constraint is specified as $D = \left\{ \left| S \cap dim_{weightloss} \right| \geq 1, \left| S \cap dim_{diet} \right| \geq 1, \left| S \cap dim_{exercise} \right| \geq 1 \right\}$, where $S$ represents the recommended challenge set, $dim_{weightloss}$ represents the weight-loss-oriented dimension, $dim_{diet}$ denotes the diet-oriented dimension, and $dim_{exercise}$ denotes the exercise-oriented dimension.

## 5 Performance Analysis and Results

We apply our recommendation framework to the weight-management context described in Section 4, with the aim of promoting users' engagement in weight-loss challenges on the platform. In particular, we implement the algorithm introduced in Section 3.2 to offer top-K challenges to users on a weekly basis.[3] The time span of recommendation is 16 weeks (i.e., the same as our data collection window). To demonstrate the effectiveness of our recommendation design, we follow the design-science paradigm to rigorously evaluate our recommendation framework through a series of experiments. We first examine the effectiveness of our deep-learning embeddings in capturing user characteristics and challenge attributes. We then apply different evaluation approaches to test each of our model components as well as to compare our model against state-of-the-art recommendation systems.

## 5.1 Experiment 1: t-SNE Visualization of Context Space

To show how the construction of deep-learning models improves feature engineering, we use t-SNE to visualize the embeddings in a two-dimensional space (Maaten and Hinton 2008). t-SNE is a nonlinear dimensionality reduction technique that is well suited for deconstructing high-dimensional data. It can project high-dimensional vectors into lower dimensions without changing the data structure so that it helps us to understand data patterns in a more intuitive way. In a t-SNE plot, two points that are close to each other indicate that the corresponding embedding representation vectors are similar. We provide the visualization results for challenge embeddings and user embeddings in Figures 3 and 4, respectively.

Specifically, we present the t-SNE plot for our constructed challenge embeddings in Figure 3(1). In Figures 3(2) and 3(3), we provide the t-SNE plots for two state-of-the-art word2vec deep-learning models, BERT and FastText. We use these two models as a benchmark for evaluating the performance of our proposed challenge-embedding model. In the plot, we use different colors to indicate each challenge type, such as weight-loss oriented, diet oriented, and exercise oriented. We use the color degree to denote the intensity level of a challenge: a deeper color indicates a challenge of higher intensity. For example, diet-oriented challenges are denoted by orange points, and among the points, there are three color degrees: light orange, dark orange, and orange-red, representing low-, medium-, and high-intensity levels, respectively. We find that, in Figure 3(1), challenges that belong to the same type are tightly clustered. In addition, within each challenge-type cluster, challenges of the same intensity level tend to be close to each other. These patterns indicate that our challenge embedding model can well capture intrinsic challenge attributes,

---

especially the ones that are key to goal-setting theory (Doran 1981). In comparison, the benchmark models cannot clearly distinguish these challenge patterns.

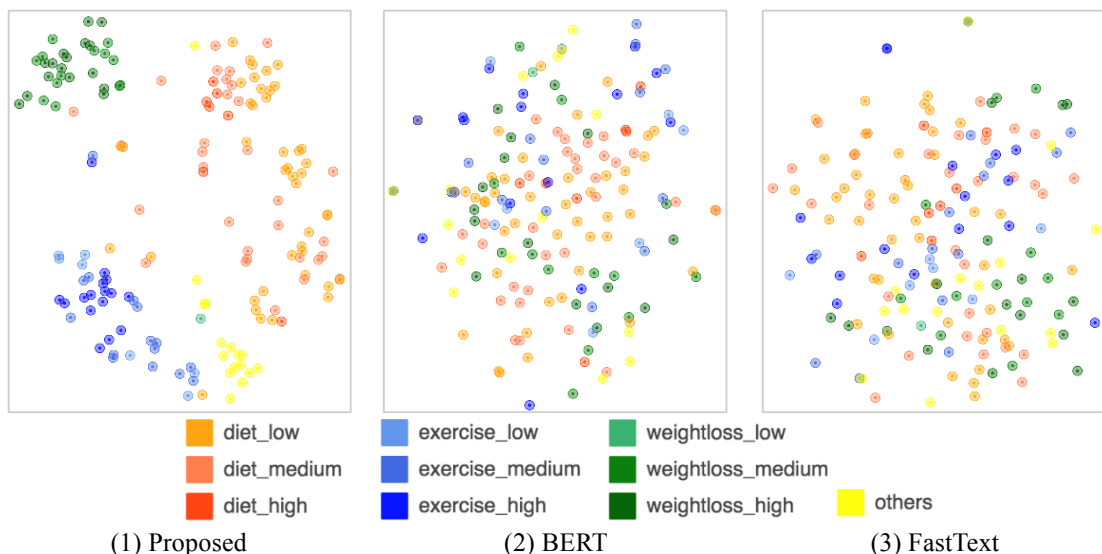

**Figure 3** Challenge Embedding t-SNE Visualization

The t-SNE plot for our constructed user embeddings is presented in Figure 4(1). As our deep-learning model incorporates users' time-varying sequence features, such as their weight curves and activity paths, each user may generate different embeddings at different times, which are denoted by different points. To show how our model captures user patterns, we use different colors to distinguish users' gender and age groups.[4] Specifically, the pink points in the plot denote female users, and the blue points denote male users; the deeper the color, the larger the age group to which users belong. In addition, we use different marker shapes to denote users' in-period weight-loss status, i.e., weight loss, weight maintenance, weight gain, and unknown status. Our visualization results show that users tend to gather into sequence clusters if they have similar features noted above. Specifically, each sequence cluster typically comprises the same gender. Users of a larger age group are typically clustered at the two sides of the plot (deeper colors), whereas users of a smaller age group generally locate in the middle (lighter colors). Finally, users whose weight remains unchanged tend to show up together, and users who have weight changes tend to gather elsewhere. These results indicate that our model can well capture user patterns in terms of gender, age, and weight-loss status. Note that these features are most directly related to our weight-loss context; in particular, gender and age are two demographics that directly affect users' body weight, and weight-loss status reflects users' in-period weight variations.

---

[4] Age groups are encoded as follows: Group 0: <=29, Group 1: 30–39, Group 2: 40–49, Group 3: 50–59, Group 4: 60–69, Group 5: >69.

We compare our proposed user-embedding model with two benchmark models, collab_learner and tabular. The results are presented in Figures 4(2) and 4(3), respectively. The collab_learner model and the tabular model are two encapsulated python learners provided in the Fastai library. In particular, the collab_learner model learns user representations from the historical challenge-selection data; however, it is not able to incorporate users' personal characteristics. The tabular model is a deep-learning model that learns user embeddings based on users' tabular attributes, such as gender and age, but does not extract signals from sequential data, such as users' health histories and behavior paths. As shown, these two models do not perform as well as our models, as there is no explicit user pattern displayed.

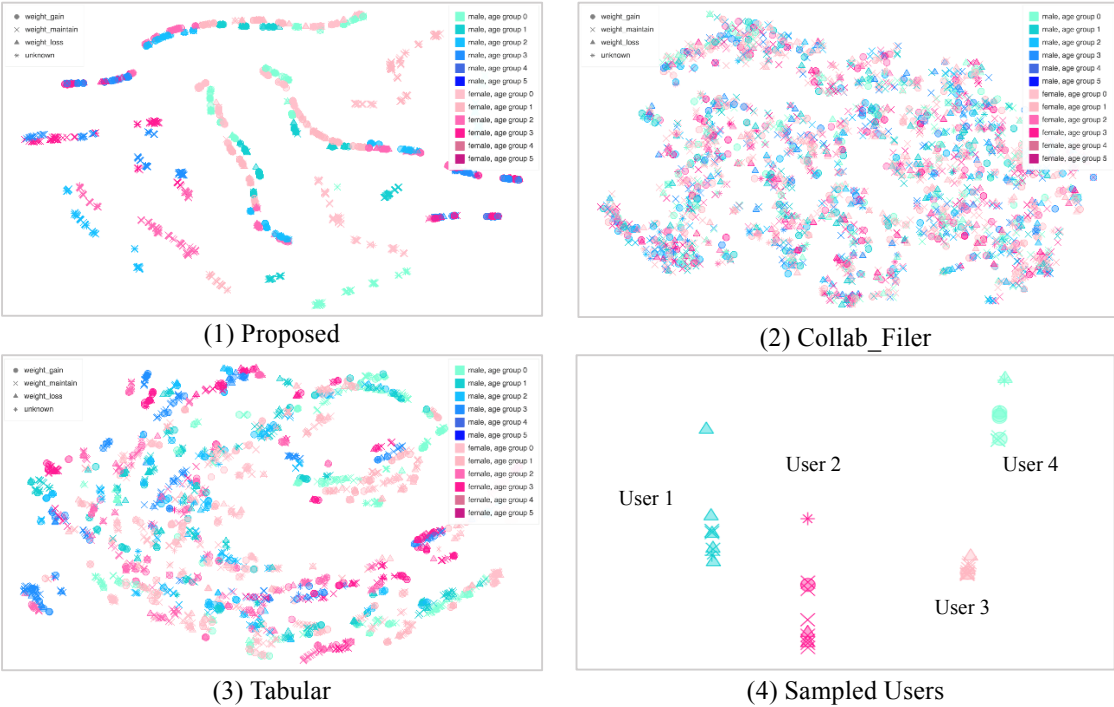

(1) Proposed          (2) Collab_Filer

(3) Tabular          (4) Sampled Users

**Figure 4** User Embedding t-SNE Visualization

Finally, the sequence shape of the user clusters produced by our model in Figure 4(1) motivates us to further investigate granular individual-level patterns, as the points in a sequence are likely generated by the same or similar users. We randomly sampled several individual users and plotted their corresponding embeddings in Figure 4(4). We find that the embeddings of the same user locate close to each other and tend to be concatenated into a trajectory and that the embeddings of different users are located relatively far apart. These results show that our proposed user-embedding model can effectively capture the sequential changes in users' behaviors. Together, Figures 3 and 4 provide evidence for the effectiveness of our deep-learning models.

## 5.2    Experiment 2: Ablation Analysis of Model Components

We conduct an ablation analysis to compare our model with a series of baseline MABs. This allows us to show that each of our model components (i.e., the deep-learning-based feature engineering procedure and the diversity constraint) is effective in helping users to find the relevant challenges. The baseline MABs are the counterpart MAB models that partially incorporate or do not incorporate the proposed model components. Specifically, the baseline MABs that we investigate include the MAB without user embeddings, the MAB without challenge embeddings, the MAB without either embeddings, the MAB without diversity constraint, and the MAB without either embeddings or constraint. When user embeddings are not incorporated, we use users' attribute features (e.g., gender, age, etc.) to account for recommendation personalization. When challenge embeddings are not incorporated, we use the annotated challenge features to capture the inherent attributes of challenges, namely, the variables listed in Table 2.

Evaluating an explore/exploit policy is difficult because we typically do not know the reward of an action that was not chosen. Possible solutions include doubly-robust estimation (Dudík et al. 2011), offline precision evaluated by preference set (Qin et al. 2014), and simulation. The first evaluation approach, doubly-robust estimation, is an offline data evaluation approach that utilizes pre-collected historical data to evaluate policy performance. The historical data are assumed to contain three sets of information: action, context, and reward. As the data are pre-collected, we are able to observe only the rewards for the chosen action in the data. To adjust for the potential bias caused by the data collection process, the doubly-robust estimator combines two policy evaluation methods, direct simulation (DS) and inverse propensity score (IPS). Formally, let $G$ denote the offline dataset, which contains action $a$, context $\mathbf{v}$, and reward $r$. In our context, action refers to the platform's provision of a challenge in the data, context $\mathbf{v}$ includes individuals' weight-management context $\mathbf{x}_{it}$ and the challenge features, and $r$ is users' feedback, that is, whether a user selects a challenge. Let $S_{it}$ denote the set of challenges recommended to user $i$ at week $t$, and $|S_{it}| = K$. Our doubly-robust estimator can be expressed as follows:

$$\bar{R} = \frac{1}{|G|} \sum_{\{i,t,a\} \in G} \left\{ \frac{1}{K} \sum_{k \in S_{it}} \left[ \hat{\varphi}_{\mathbf{v}_{ak}} + \frac{(r_{a,\mathbf{x}_{it}} - \hat{\varphi}_{\mathbf{v}_{iak}}) I_{\{k=a\}}}{\hat{p}(a \mid \mathbf{x}_{it})} \right] \right\}, \tag{4}$$

where $\hat{\varphi}$ is a reward simulator, and $\hat{p}$ is the propensity of challenge provision in the data. The rationale of this method is that, when data are not available, the method uses a pre-trained reward predictor to simulate the reward; otherwise, it applies a correction to the reward predictor using the actual data.

The second evaluation method measures recommendation precision. Following previous studies (Qin et al. 2014; Qin and Zhu 2013), we construct each user's preference set as the set of challenges selected by the user in the data. The recommendation precision is thus the overlap ratio between the recommendation set and the preference set. Finally, in light of previous theoretical MAB studies (Hertz et al. 2018; Sani et al. 2012), we examine our model performance through a simulated environment. Specifically, we construct a logistic predictor for users' binary challenge-selection decisions, that is, $r_t(i, k) = (1 + \exp(-\mathbf{v}_{itk}^T \zeta))^{-1}$, where $\mathbf{v}$ is a concatenation of user embeddings and challenge embeddings. The weight vector $\zeta$ could have been chosen arbitrarily, but it was in fact a perturbed version of the weight vector trained on a randomly constructed training set (Nguyen et al. 2017), and the performance evaluation is conducted on a test set. This simulator is omniscient, in the sense of full knowledge of users' preferences and the actual amount of reward accrued by recommendations. We provide the details of these evaluation approaches in Appendix A6.

The evaluation results are provided in Table 3. Each value in the table represents users' average selection rate during the entire recommendation course. A superscripted asterisk denotes that a benchmark model performs significantly worse than the proposed model. Our results show that the MABs that include only one of the components have an inferior performance (i.e., lower average challenge-selection rate) than our proposed model. Specifically, the MAB without diversity constraint is suggested to be significantly worse by the doubly-robust estimation and simulation method. The MAB without user embeddings or challenge embeddings (or both) is shown to have a worse performance by all three evaluation methods. Finally, we find that the MAB without either embeddings or diversity constraint performs worse than the MABs that partially incorporate the model components. These results indicate that each of our proposed model components is effective. As compared to users' attribute features, user embeddings can better capture the sequential information embedded in users' health histories and behavior paths and, thus, are more effective. Challenge embeddings are more effective than the annotated challenge features, as they are able to extract semantic information from the textual challenge descriptions. In addition, as most annotated challenge features are categorical and one-hot encoded, they may not provide much information for the learning process. In comparison, challenge embeddings can better calibrate the similarity among challenges and make the learning more effective.

**Table 3** Comparison of Baseline MABs

| Models | Doubly-Robust Estimation | Offline Precision | Omniscient Simulator |
|---|---|---|---|
| proposed MAB | 0.5115 | 0.6054 | 0.4467 |
| *Baseline MABs* | | | |
| No user embeddings | 0.4983 ** | 0.5933 * | 0.4275 ** |
| No challenge embeddings | 0.4722 *** | 0.5625 *** | 0.4215 *** |
| No embeddings | 0.4685 * | 0.5519 *** | 0.4206 *** |
| No diversity constraints | 0.5028 *** | 0.5974 | 0.4394 * |
| No embeddings, no constraints | 0.4583 *** | 0.5462 *** | 0.4172 ** |

Note: Asterisk in superscript denotes that a benchmark model performs significantly worse than the proposed model. Significance levels are: $^*\, p < 0.1$, $^{**}\, p < 0.05$, $^{***}\, p < 0.01$.

In Figure 5, we plot the recommendation performance across time to show the learning curve of each model. The $x$-axis denotes recommendation rounds, and the $y$-axis denotes the average challenge selection rate up to round $t$. It is shown that our model achieves the highest learning rate across all periods, regardless of the evaluation approach taken. That is, our model can boost the average challenge-selection rate faster than the benchmark models can. For example, when evaluated by simulation, our model increases the average challenge selection rate from approximately 0.39 to approximately 0.45 after the 16-week recommendation phase, which is a 15% increase. This is followed by the MAB with no diversity constraint (~13%), the MAB with no user embeddings (~11%), the MAB with no challenge embeddings (~10%), the MAB without either embedding (~8%), and the MAB without either diversity constraint or embeddings (~8%). These results further highlight the effectiveness of our deep-learning-based feature engineering and diversity constraint in the learning procedure.

### 5.3 Experiment 3: Comparison with Benchmark Recommendation Systems

We compare our proposed recommendation framework against a wide range of benchmark models. In particular, for benchmark bandit models, we consider UCB and $\varepsilon$-greedy, which are two classic online-learning methods to solve the "exploitation-versus-exploration" tradeoff. For batch-learning-based models, we consider a variety of collaborative filtering methods, such as context-aware recommenders and matrix-factorization-based models. We also consider content-based filtering and hybrid filtering. Content-based filtering is able to offer recommendations based on item features. Hybrid filtering further combines content-based filtering with collaborative filtering to incorporate information embedded in users' challenge selection histories. As the batch-learning-based models make recommendations by exploiting users' item-selection histories, we use the first four weeks of data to train the algorithms. Finally, we consider pure exploitation and pure exploration, which are two recommendation schemes that do not seek a balance between exploitation and exploration. We summarize our benchmark models in Table 4. The implementation details of these benchmark models are provided in Appendix A7. Similarly, we calibrate recommendation performance by users' average challenge-selection rate and evaluate it through the aforementioned three evaluation approaches, i.e., doubly-robust estimation, offline precision, and simulation. Our evaluation results are provided in Table 5.

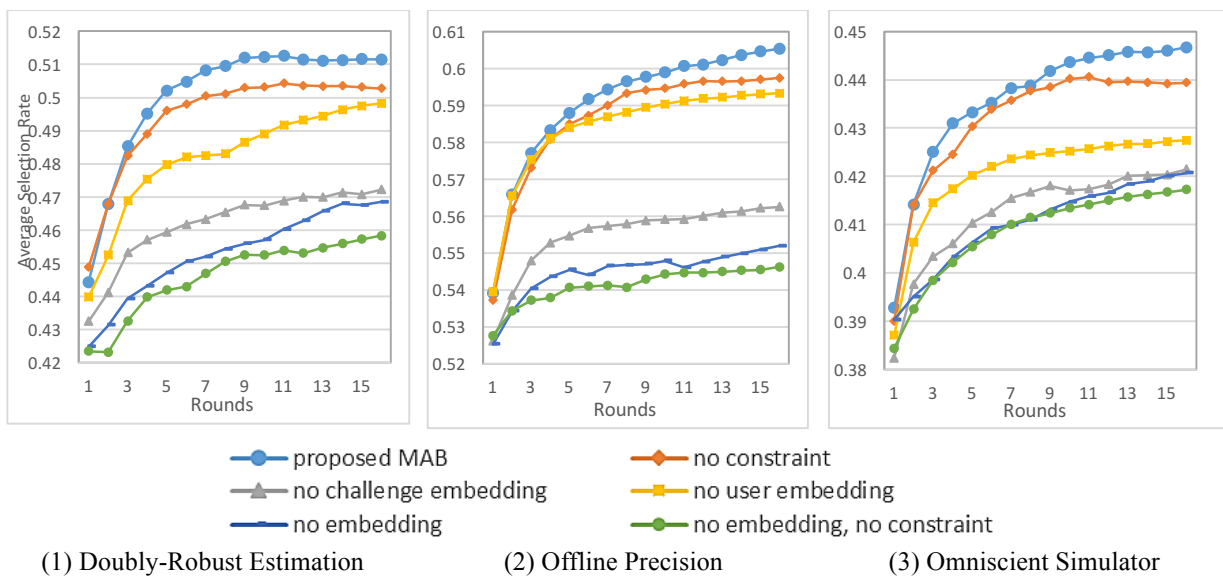

| | |
|---|---|
| ● proposed MAB | ● no constraint |
| ▲ no challenge embedding | ■ no user embedding |
| ● no embedding | ● no embedding, no constraint |

(1) Doubly-Robust Estimation    (2) Offline Precision    (3) Omniscient Simulator

**Figure 5** The Learning Curves of Our Model and the Baseline MABs

The results show that our proposed model has the best recommendation performance under all evaluation measures. It achieves an average challenge selection rate of 51.15% when evaluated by doubly-robust estimation; 60.54%, by offline precision; and 44.67%, by simulation. UCB and $\varepsilon$-greedy are shown to have comparable performance (~42% for doubly-robust estimation, ~52% for offline precision, and ~39% for simulation), but both perform significantly worse than our model. Batch-learning-based models generally do not perform well (mainly below 40%). The performance inferiority may be due to users' dynamic preferences for weight-loss challenges. In other words, the batch-learning-based models assume that users' preferences can be well represented by their past behavior patterns (Adomavicius and Tuzhilin 2005; Sahoo et al. 2012); thus, these models can be biased when users' preferences contain dynamic variations that are not fully captured by the historical behavior data.

**Table 4** Summary of Comparison Benchmark Recommendation Algorithms

| Benchmark Models | Recommendation Algorithm Description |
|---|---|
| UCB | A bandit algorithm that incorporates the opportunity for exploration by choosing the arm with the highest upper bound of the reward confidence interval; |
| $\varepsilon$ -greedy | A bandit algorithm that chooses the arm with the seemingly highest average reward with probability $1-\varepsilon$ and explores a random arm with probability $\varepsilon$ ; |
| CACF | Context-Aware Collaborative Filtering, which incorporates the contexts of users' item selections as weights into a normal collaborative filtering procedure (Chen 2005); |
| SCF | Social Collaborative Filtering, which formulates a neighborhood-based method for cold-start collaborative filtering in a generalized matrix algebra framework (Sedhain et al. 2014); |
| PMF | Probabilistic Matrix Factorization, a model-based collaborative filtering approach that uses matrix factorization under a probabilistic framework to estimate user-item interactions (Mnih and Salakhutdinov 2008); |
| CAMF | Context-Aware Matrix Factorization, which is an extension of the classic matrix factorization approach for incorporating contextual information (Baltrunas et al. 2011); |
| CB | Content-based filtering, an approach to offer recommendations based on content similarities of items (Bieliková et al. 2012; Pon et al. 2007); |
| hybrid_pure | A hybrid model that combines pure collaborative filtering with CB using mixed hybridization (Burke 2002); |
| hybrid_cacf | A hybrid model that combines CACF and CB using mixed hybridization; |
| pure exploitation | A model that selects the best option given current knowledge; |
| pure exploration | A model that fully randomizes recommendations. |

Finally, our results show that pure exploitation and pure exploration achieve worse recommendation performance as compared to our model. This performance gap indicates the importance and necessity of balancing the exploitation-versus-exploration tradeoff. A pure exploitation method may be stuck in a worse local optimum. Pure exploration, in contrast, over-explores users' preferences; it fully randomizes the recommendations without utilizing or learning from information embedded in users' past behaviors. Note that pure exploitation and pure exploration can often be seen in the design of A/B testing. Specifically, in an A/B test, experimenters first spend a short time period for pure exploration, whereby they randomly assign users to different groups to examine the performance of policy variants. They then engage in a long period of pure exploitation, assigning all of the users to the group that achieves the best performance. In practice, the pure exploratory phase can be expensive or even infeasible to implement. For example, in a health-management context, it is usually infeasible to arbitrarily assign individuals to a treatment plan. Instead of two distinct periods of pure exploration and pure exploitation, a bandit-driven design adaptively combines exploration and exploitation. Thus, it can reduce the opportunity cost incurred in the exploratory phase and help service providers to achieve better performance.

**Table 5** Comparison with State-of-the-Art Recommendation Systems

| Models | Doubly-Robust Estimation | Offline Precision | Omniscient Simulator |
|---|---|---|---|
| proposed MAB | 0.5115 | 0.6054 | 0.4467 |
| *Baseline State-of-the-Art Bandit Models* | | | |
| UCB | 0.4199 *** | 0.5266 *** | 0.3981 *** |
| $\varepsilon$ -greedy | 0.4193 *** | 0.5242 *** | 0.3848 *** |
| *Baseline State-of-the-Art Batch-Learning Models* | | | |
| CACF | 0.4294 *** | 0.3149 *** | 0.4325 ** |
| SCF | 0.3696 *** | 0.3970 *** | 0.4005 *** |
| PMF | 0.3618 *** | 0.2124 *** | 0.3943 *** |
| CAMF | 0.3100 *** | 0.2653 *** | 0.2824 *** |
| CB | 0.4015 *** | 0.2660 *** | 0.4330 ** |
| hybrid_pure | 0.2951 *** | 0.2492 *** | 0.3046 *** |
| hybrid_cacf | 0.4109 *** | 0.2897 *** | 0.4338 ** |
| *Pure Exploitation and Pure Exploration* | | | |
| pure exploitation | 0.4785 ** | 0.5760 *** | 0.4188 *** |
| pure exploration | 0.4617 *** | 0.5690 *** | 0.4006 *** |

Note: Asterisk in superscript denotes that a benchmark model performs significantly worse than the proposed model. Significance levels are: $^{*} p < 0.1$, $^{**} p < 0.05$, $^{***} p < 0.01$.

## 5.4 Experiment 4: On Dynamics of Users' Preferences

In a typical healthcare context, individuals' behavior patterns likely change over time (Johnson et al. 2002; King et al. 2006). In this experiment, we investigate the recommendation results for the users whose choices tend to vary considerably as a means to examine whether our recommendation framework can well capture the dynamics in users' preferences. From the test-user set, we select the 30 users whose challenge choices have the largest embedding variance.[5] We implement our recommendation algorithm for the selected users and compare the recommendation performance with that of using the full test-user set. Table 6 presents the results for the new test-user set. Our model is shown to outperform all of the benchmark models by all evaluation measurements. In particular, our model achieves an average challenge selection rate of 53.44% when evaluated by doubly-robust estimation; 60.90%, by offline precision; and 47.56%, by simulation.

To better show the performance variations, we plot the differences between the new recommendation results and the original results on the full test set in Figure 6. Here, we present the performance changes measured by doubly-robust estimation. The performance changes measured by offline precision and simulation are similar and are provided in Appendix A8. We find that bandit-driven models, such as our proposed model, UCB, and $\varepsilon$-greedy, have different degrees of performance increase. In contrast, batch-learning-based models generally have a performance decrease. These results suggest that the advantage of the online-learning scheme is further strengthened when evaluated on the dynamic users whose preferences tend to vary frequently. The performance decrease of batch-learning-based models indicates that the "first learn, then earn" recommendation scheme may perform even worse when users' preferences exhibit strong dynamics. This may be because the dynamic patterns in users' preferences cannot be fully captured by the historical data. Bandit-driven models, in contrast, are able to actively collect users' feedback on recommendations and, thus, can capture changes in users' preferences more promptly.

---

[5] As challenge embeddings are vectors, we define the variance of embeddings as the minimum value of the variances of the elements.

**Table 6** Results for Dynamic Test Users

| Method | Doubly-Robust Estimation | Offline Precision | Omniscient Simulator |
|---|---|---|---|
| proposed MAB | 0.5344 | 0.6090 | 0.4756 |
| *Benchmark Models* | | | |
| UCB | 0.4287 *** | 0.5266 *** | 0.3981 *** |
| epsilon greedy | 0.4415 *** | 0.5242 *** | 0.3848 *** |
| CACF | 0.4183 *** | 0.2278 *** | 0.4378 *** |
| SCF | 0.3524 *** | 0.3964 *** | 0.3783 *** |
| PMF | 0.3494 *** | 0.1975 *** | 0.3894 *** |
| CAMF | 0.3061 *** | 0.2567 *** | 0.2800 *** |
| CB | 0.3694 *** | 0.2317 *** | 0.4350 *** |
| hybrid_pure | 0.2772 *** | 0.2208 *** | 0.3005 *** |
| hybrid_cacf | 0.3942 *** | 0.2250 *** | 0.4308 *** |

Note: Asterisk in superscript denotes that a benchmark model performs significantly worse than the proposed model. Significance levels are: $^{*} p < 0.1$, $^{**} p < 0.05$, $^{***} p < 0.01$.

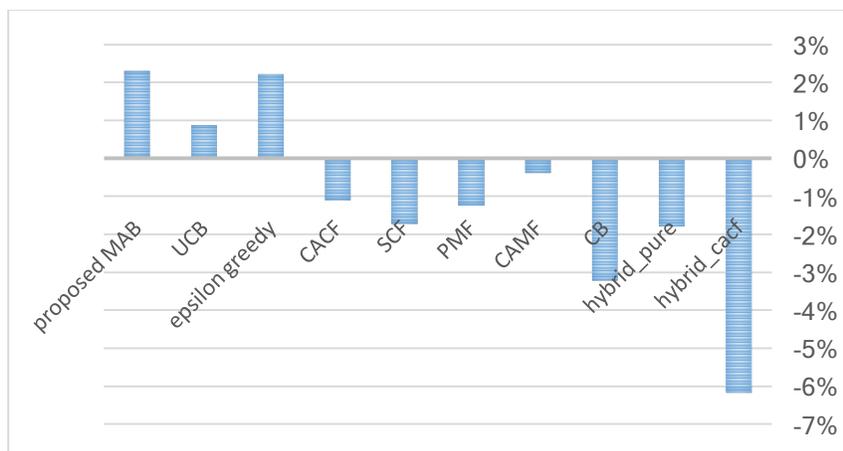

**Figure 6** Performance Variation for Dynamic Users

## 5.5 Experiment 5: Diversity Analysis

We evaluate the recommendation diversity across challenge types to examine whether our model can effectively learn users' diverse challenge preferences in the data. Specifically, for our proposed model and each of the benchmark models, we calculate the recommendation frequency for each challenge type to construct a diversity distribution. We then compare the recommendation frequencies with users' challenge selection frequencies in the data, which reveal users' true preferences of challenge types. We use the Jensen-Shannon divergence (JSD) to measure the similarity between two diversity distributions (Endres and Schindelin 2003; Fuglede and Topsoe 2004). A small JSD value indicates high similarity between two diversity distributions.

In Figure 7, we visualize the diversity distribution for each recommendation model and present the corresponding JSD values. The first bar in the figure represents the diversity distribution in users' challenge-selection histories observed in the data. The second bar represents the diversity distribution in the recommendations provided by our proposed model. As can be seen, the first two bars are very similar to each other, indicating that the recommendations produced by our model are diversified in a way that is similar to users' actual challenge-selection histories. The recommendation diversity distributions produced by the benchmark models generally have a larger difference from the observed challenge-selection data. For example, the bar that corresponds to CACF is quite different from the first bar. These observations are confirmed by our JSD results, that our proposed model has the smallest JSD value (i.e., 0.03) across all of the benchmark models. Combined with the results in the earlier-discussed experiments, these findings provide evidence that our proposed recommendation framework can well support users' diverse healthcare preferences and that the diversity constraint can further guide the recommendation system to explore along each of the weight-management dimensions and, thus, improves learning efficiency.

## 5.6   Experiment 6: User Improvement

In this experiment, we aim to examine whether our recommendation framework can benefit more users. We define user improvement as the percentage of users who receive more preferred items from a focal recommendation algorithm than from a baseline algorithm. We use probabilistic matrix factorization (PMF) as our baseline algorithm. PMF models hidden user representations based on the users' challenge-selection histories. It does not, however, incorporate contextual information of users' selection behaviors, and it is batch-learning-based. Thus, by comparing with PMF, we are able to assess the value of the recommendation context along with the online-learning scheme. We present our results for user improvement in Figure 8. It is shown that our proposed recommendation approach has the highest user improvement rate (~76.19%), suggesting that approximately 76% of users receive more preferred challenges from our recommendation framework than from PMF. Comparatively, the other recommendation approaches have a lower user improvement rate (all below 70%). These results indicate that our recommendation framework can serve to improve a larger user population on the platform.

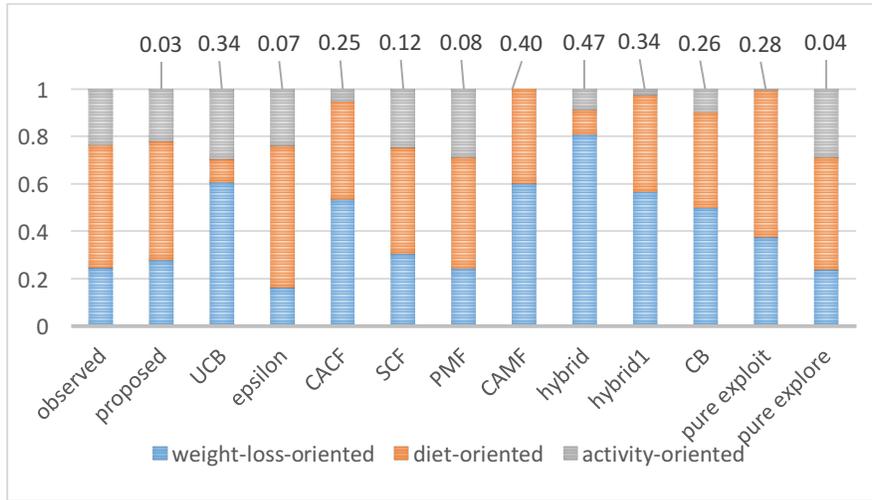

**Figure 7** Diversity Distribution

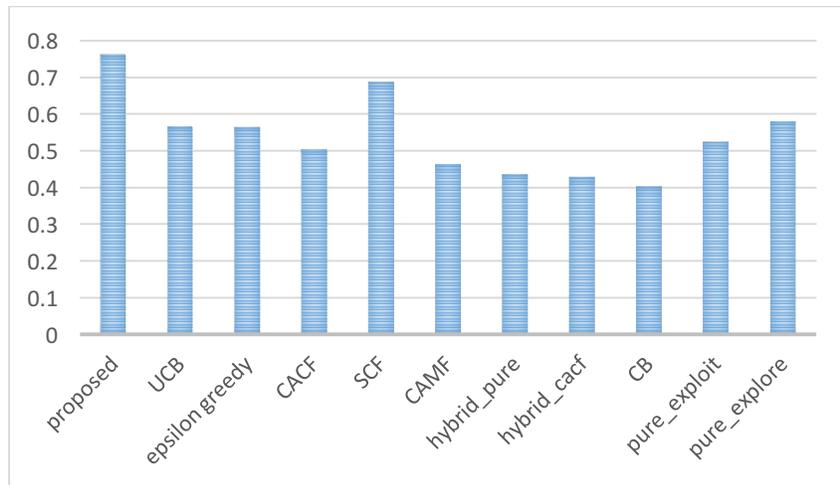

**Figure 8** User Improvement: Using PMF as Baseline

## 5.7 Experiment 7: Using Our Recommendation Framework to Improve Weight Loss

The former experiments evaluate the effectiveness of our recommendation framework in improving users' challenge-selection rates. In the healthcare context, it is also important that service providers take further steps to evaluate corresponding health-related outcomes. Although necessary, increasing users' engagement in interventions may not be enough, as it does not directly guarantee an improvement in health. Therefore, in this experiment, we further examine our recommendation performance in improving users' weight-loss outcomes. In particular, we explore a different recommendation target: users' in-period weight-loss rate.[6] That is, we use users' in-period weight-loss rate as feedback to guide the learning of our recommendation framework. Different from prior experiments, generating weight-loss feedback requires two steps of simulation. First, we use the logistic predictor described in Section 5.2 to simulate whether users will choose a particular item. Second, we simulate users' in-period weight-loss status (e.g., weight gain or non-gain) based on their choices. We use users' weigh-in data to train a logistic predictor for this simulation, which takes user embeddings and the average challenge embeddings of users' choice as the input variables and users' weight-loss status as the prediction target.

As shown in Table 7, our proposed recommendation framework has the best performance under the new recommendation target, achieving an average in-period weight-loss rate of 75.64%. In contrast, the benchmark models achieve a significantly worse average in-period weight-loss rate. These results indicate that our recommendation design is effective not only in helping users to find preferable challenges to engage in but also in further helping them to achieve better weight-loss performance.

**Table 7** Weight Loss as Recommendation Goal

| Method | Average In-Period Weight-loss Rate |
|---|---|
| proposed MAB | 0.7564 |
| *State-of-the-Art Bandit Models* | |
| UCB | 0.7039 *** |
| epsilon greedy | 0.6739 *** |
| *State-of-the-Art Batch-Learning Models* | |
| CACF | 0.6318 *** |
| SCF | 0.6294 *** |
| PMF | 0.6459 *** |
| CAMF | 0.6165 *** |
| CB | 0.6472 *** |
| hybrid_pure | 0.6049 *** |
| hybrid_cacf | 0.6507 *** |
| *Pure exploitation and Pure Exploration* | |
| pure exploitation | 0.7035 *** |
| pure exploration | 0.4006 *** |

Note: Asterisk in superscript denotes that a benchmark model performs significantly worse than the proposed model. Significance levels are: $^*\,p < 0.1$, $^{**}\,p < 0.05$, $^{***}\,p < 0.01$.

---

[6] In this study, we define in-period weight-loss rate as the percentage of periods in which users' weight has decreased or remained unchanged.

# 6    Conclusion

In this study, we take a design-science perspective to develop a novel recommendation framework for providing personalized healthcare recommendations to users on online healthcare platforms. The design of our recommendation framework is motivated by several unique patterns in individuals' health behaviors. First, due to the evolving process of health management, users' health behaviors may continuously change with time. Second, users may need multiplex healthcare information to manage different health aspects. These characteristics indicate that users' healthcare preferences can be dynamic and diverse. To this end, we propose a deep-learning and diversity-enhanced MAB framework. MAB is able to adaptively learn users' changing behavior patterns while promoting diversity along the exploration process. To better adapt an MAB to the healthcare recommendation context, we further synthesize two model components into our framework based on prominent health-behavior theories. The first component is a deep-learning-based feature construction procedure, aimed at capturing important healthcare recommendation contexts, such as healthcare intervention's inherent attributes and individuals' evolving health condition and health-behavior paths. The second component is a diversity constraint, which we use to ensure that recommendations are provided in each of the major health-management dimensions, so that individuals can receive well-rounded support for their health management. We conduct a series of experiments to test each of our model components as well as to compare our model against state-of-the-art recommendation systems, using data collected from a representative online weight-loss platform. The results of the experiments provide strong evidence for the effectiveness of our proposed recommendation framework.

Our study contributes to the emerging literature on the application of business intelligence (Abbasi et al. 2016; Chen et al. 2012). We demonstrate that prescriptive analytics can be integrated with IT artifacts to generate applicable insights. In particular, the innovative healthcare recommendation framework that we developed provides an important contribution to the literature on recommendation systems and online healthcare systems. To the best of our knowledge, we are among the first to combine MAB models with deep-learning-based embeddings to improve the characterization of recommendation contexts. In addition, the inclusion of diversity constraints demonstrates a way of promoting recommendation diversity according to pre-designed dimensions. This innovation can be of significance in professional industries, in which domain expertise needs to be incorporated to guide recommendation diversification.

From a practical perspective, our recommendation framework can be used to address real-world challenges in healthcare recommendation problems. The effectiveness of our framework, as demonstrated by our results, implies great potential for using our recommendation design to provide users with tailored engagement suggestions. Although our recommendation design is proposed for assisting individuals' engagement in online health management, the framework can be extended to broader problem settings. For example, the combination of an MAB and deep-learning-based feature engineering can be used to solve other healthcare problems, such as drug discovery, disease diagnosis, clinical trials, and therapy development. Decision making in such healthcare problems usually involves complex contextual knowledge (e.g., drug structures, patient health histories, symptom development paths), and decision-makers usually do not have full knowledge of the environment (e.g., whether a drug is effective). The online-learning framework of an MAB can help decision-makers to better cope with uncertainties in the healthcare environment, and deep-learning models can be combined to improve the characterization of decision-making contexts. In addition, we demonstrate a way to formulate recommendation constraints, which can be used to incorporate domain expertise to guide the recommendation procedure.

Our recommendation framework can also be extended to fields beyond healthcare. Real-world decision-making problems, such as financial investment, product pricing, and marketing, usually contain different levels of uncertainty. The uncertainty may be because decision-makers do not gather enough data to guide their decision making, or the decision-making environment is frequently changing (e.g., market instability, technology change, policy environment fluctuation). Thus, it is of practical importance to develop an adaptive decision-making framework that can respond well to the uncertainty in the environment. Decision-makers may consider combining an MAB with deep-learning embeddings to learn the context-dependency of their decision results while adaptively adjusting their strategies to minimize opportunity cost. In addition, in real-world recommendation problems, it is usually desirable to recommend diversified content to maximize the coverage of the information that users find interesting to improve their engagement experience. Our formulation of the diversity constraint can be used to strengthen recommendation diversification along with theory-guided dimensions in multiple application areas.

# References

Abbasi A, Sarker S, and Chiang R H. (2016) Big Data Research in Information Systems: Toward an Inclusive Research Agenda. *Journal of the Association for Information Systems* 17(2): 3.

Adomavicius G, and Tuzhilin A. (2005) Toward the Next Generation of Recommender Systems: A Survey of the State-of-the-Art and Possible Extensions. *IEEE Transactions on Knowledge and Data Engineering* 17(6): 734-749.

Auer P, Cesa-Bianchi N, and Fischer P. (2002) Finite-Time Analysis of the Multiarmed Bandit Problem. *Machine Learning* 47(2-3): 235-256.

Baltrunas L, Ludwig B, and Ricci F. (2011) Matrix Factorization Techniques for Context Aware Recommendation. *Proceedings of the Fifth ACM Conference on Recommender Systems*: 301-304.

Bandura A. (1991) Social Cognitive Theory of Self-Regulation. *Organizational Behavior and Human Decision Processes* 50(2): 248-287.

Bandura A. (1998) Health Promotion from the Perspective of Social Cognitive Theory. *Psychology and Health* 13(4): 623-649.

Bieliková M, Kompan M, and Zeleník D. (2012) Effective Hierarchical Vector-Based News Representation for Personalized Recommendation. *Computer Science and Information Systems* 9(1): 303-322.

Burke R. (2002) Hybrid Recommender Systems: Survey and Experiments. *User Modeling and User-Adapted Interaction* 12(4): 331-370.

CDC. (2019) About Chronic Diseases, Accessed Oct 23, 2019. *https://www.cdc.gov/chronicdisease/about/index.htm#:~:text=Many%20chronic%20diseases%20are%20caused,Lack%20of%20physical%20activity*.

CDC. (2020) People with Certain Medical Conditions, Accessed Aug 10, 2020. *https://www.cdc.gov/coronavirus/2019-ncov/need-extra-precautions/people-with-medical-conditions.html#obesity*.

Chapelle O, and Li L (2011) An Empirical Evaluation of Thompson Sampling. *Advances in neural information processing systems*), 2249-2257.

Chen A (2005) Context-Aware Collaborative Filtering System: Predicting the User's Preference in the Ubiquitous Computing Environment. *International Symposium on Location-and Context-Awareness* (Springer), 244-253.

Chen H, Chiang R H, and Storey V C. (2012) Business Intelligence and Analytics: From Big Data to Big Impact. *MIS Quarterly*: 1165-1188.

Cheng H-T, Koc L, Harmsen J, Shaked T, Chandra T, Aradhye H, Anderson G, Corrado G, Chai W, and Ispir M. (2016) Wide & Deep Learning for Recommender Systems. *Proceedings of the 1st Workshop on Deep Learning for Recommender Systems*: 7-10.

Cohen J D, McClure S M, and Yu A J. (2007) Should I Stay or Should I Go? How the Human Brain Manages the Trade-Off between Exploitation and Exploration. *Philosophical Transactions of the Royal Society B: Biological Sciences* 362(1481): 933-942.

DiMatteo M R. (2004) Social Support and Patient Adherence to Medical Treatment: A Meta-Analysis. *Health Psychology* 23(2): 207.

Doran G T. (1981) There'sa Smart Way to Write Management's Goals and Objectives. *Management Review* 70(11): 35-36.

Dudík M, Langford J, and Li L. (2011) Doubly Robust Policy Evaluation and Learning. *arXiv preprint arXiv:1103.4601*.

Endres D M, and Schindelin J E. (2003) A New Metric for Probability Distributions. *IEEE Transactions on Information Theory* 49(7): 1858-1860.

Fleder D, and Hosanagar K. (2009) Blockbuster Culture's Next Rise or Fall: The Impact of Recommender Systems on Sales Diversity. *Management Science* 55(5): 697-712.

Franz M J. (2001) The Answer to Weight Loss Is Easy—Doing It Is Hard! *Clinical Diabetes* 19(3): 105-109.

Fuglede B, and Topsoe F (2004) Jensen-Shannon Divergence and Hilbert Space Embedding. *Proceedings of International Symposium on Information Theory.* (IEEE), 31.

Gittins J C. (1979) Bandit Processes and Dynamic Allocation Indices. *Journal of the Royal Statistical Society: Series B (Methodological)* 41(2): 148-164.

Greenewald K, Tewari A, Murphy S, and Klasnja P. (2017) Action Centered Contextual Bandits. *Advances in Neural Information Processing Systems*: 5977-5985.

Hall K D, and Kahan S. (2018) Maintenance of Lost Weight and Long-Term Management of Obesity. *Medical Clinics of North America* 102(1): 183-197.

Hertz U, Bahrami B, and Keramati M. (2018) Stochastic Satisficing Account of Confidence in Uncertain Value-Based Decisions. *PLOS One* 13(4): e0195399.

Isinkaye F, Folajimi Y, and Ojokoh B. (2015) Recommendation Systems: Principles, Methods and Evaluation. *Egyptian Informatics Journal* 16(3): 261-273.

Johnson F, and Wardle J. (2011) The Association between Weight Loss and Engagement with a Web-Based Food and Exercise Diary in a Commercial Weight Loss Programme: A Retrospective Analysis. *International Journal of Behavioral Nutrition and Physical Activity* 8(1): 83.

Johnson P E, Veazie P J, Kochevar L, O'connor P J, Potthoff S J, Verma D, and Dutta P. (2002) Understanding Variation in Chronic Disease Outcomes. *Health Care Management Science* 5(3): 175-189.

Kim M J, and Lim A E. (2015) Robust Multiarmed Bandit Problems. *Management Science* 62(1): 264-285.

King G, Willoughby C, Specht J A, and Brown E. (2006) Social Support Processes and the Adaptation of Individuals with Chronic Disabilities. *Qualitative Health Research* 16(7): 902-925.

Konstan J A, and Riedl J. (2012) Recommender Systems: From Algorithms to User Experience. *User Modeling and User-Adapted Interaction* 22(1-2): 101-123.

Krukowski R A, Harvey-Berino J, Ashikaga T, Thomas C S, and Micco N. (2008) Internet-Based Weight Control: The Relationship between Web Features and Weight Loss. *Telemedicine and e-Health* 14(8): 775-782.

Les MacLeod EdD M. (2012) Making Smart Goals Smarter. *Physician Executive* 38(2): 68.

Li L, Chu W, Langford J, and Schapire R E (2010) A Contextual-Bandit Approach to Personalized News Article Recommendation. *Proceedings of the 19th International Conference on World Wide Web* (ACM), 661-670.

Liu Y, Zhang Y, Wu Q, Miao C, Cui L, Zhao B, Zhao Y, and Guan L. (2019) Diversity-Promoting Deep Reinforcement Learning for Interactive Recommendation. *arXiv preprint arXiv:1903.07826*.

Locke E A, and Latham G P (1990) *A Theory of Goal Setting & Task Performance* (Prentice-Hall, Inc.),

Maaten L v d, and Hinton G. (2008) Visualizing Data Using T-Sne. *Journal of Machine Learning Research* 9(Nov): 2579-2605.

Mehlhorn K, Newell B R, Todd P M, Lee M D, Morgan K, Braithwaite V A, Hausmann D, Fiedler K, and Gonzalez C. (2015) Unpacking the Exploration–Exploitation Tradeoff: A Synthesis of Human and Animal Literatures. *Decision* 2(3): 191.

Misra K, Schwartz E M, and Abernethy J. (2019) Dynamic Online Pricing with Incomplete Information Using Multiarmed Bandit Experiments. *Marketing Science*.

Mnih A, and Salakhutdinov R R. (2008) Probabilistic Matrix Factorization. *Advances in Neural Information Processing Systems*: 1257-1264.

Nguyen K, Daumé III H, and Boyd-Graber J. (2017) Reinforcement Learning for Bandit Neural Machine Translation with Simulated Human Feedback. *arXiv preprint arXiv:1707.07402*.

Nutting P A, Crabtree B F, Miller W L, Stange K C, Stewart E, and Jaén C. (2011) Transforming Physician Practices to Patient-Centered Medical Homes: Lessons from the National Demonstration Project. *Health Affairs* 30(3): 439-445.

Ogbeiwi O. (2017) Why Written Objectives Need to Be Really Smart. *British Journal of Healthcare Management* 23(7): 324-336.

Ogbeiwi O. (2018) General Concepts of Goals and Goal-Setting in Healthcare: A Narrative Review. *Journal of Management & Organization*: 1-18.

Oulasvirta A, Hukkinen J P, and Schwartz B. (2009) When More Is Less: The Paradox of Choice in Search Engine Use. *Proceedings of the 32nd International ACM SIGIR Conference on Research and Development in Information Retrieval*: 516-523.

Pariser E (2011) *The Filter Bubble: How the New Personalized Web Is Changing What We Read and How We Think* (Penguin, UK)

Pathak B, Garfinkel R, Gopal R D, Venkatesan R, and Yin F. (2010) Empirical Analysis of the Impact of Recommender Systems on Sales. *Journal of Management Information Systems* 27(2): 159-188.

Pew Research Center. (2020) From Virtual Parties to Ordering Food, How Americans Are Using the Internet During Covid-19, Accessed August 31, 2020. *https://www.pewresearch.org/fact-tank/2020/04/30/from-virtual-parties-to-ordering-food-how-americans-are-using-the-internet-during-covid-19/*.

Pon R K, Cardenas A F, Buttler D, and Critchlow T (2007) Tracking Multiple Topics for Finding Interesting Articles. *Proceedings of the 13th ACM SIGKDD International Conference on Knowledge Discovery and Data Mining*), 560-569.

Pu P, Chen L, and Hu R (2011) A User-Centric Evaluation Framework for Recommender Systems. *Proceedings of the Fifth ACM Conference on Recommender Systems* (ACM), 157-164.

Qin L, Chen S, and Zhu X (2014) Contextual Combinatorial Bandit and Its Application on Diversified Online Recommendation. *Proceedings of the 2014 SIAM International Conference on Data Mining* (SIAM), 461-469.

Qin L, and Zhu X (2013) Promoting Diversity in Recommendation by Entropy Regularizer. *Twenty-Third International Joint Conference on Artificial Intelligence*.

Ricci F, Rokach L, and Shapira B (2015) Recommender Systems: Introduction and Challenges. *Recommender Systems Handbook* (Springer),1-34.

Sahoo N, Singh P V, and Mukhopadhyay T. (2012) A Hidden Markov Model for Collaborative Filtering. *MIS Quarterly*: 1329-1356.

Sani A, Lazaric A, and Munos R. (2012) Risk-Aversion in Multi-Armed Bandits. *Advances in Neural Information Processing Systems*: 3275-3283.

Schwartz E M, Bradlow E T, and Fader P S. (2017) Customer Acquisition Via Display Advertising Using Multi-Armed Bandit Experiments. *Marketing Science* 36(4): 500-522.

Sedhain S, Sanner S, Braziunas D, Xie L, and Christensen J. (2014) Social Collaborative Filtering for Cold-Start Recommendations. *Proceedings of the 8th ACM Conference on Recommender Systems*: 345-348.

Shumaker S A, and Brownell A. (1984) Toward a Theory of Social Support: Closing Conceptual Gaps. *Journal of Social Issues* 40(4): 11-36.

Snyderman R, and Dinan M A. (2010) Improving Health by Taking It Personally. *Jama* 303(4): 363-364.

Speekenbrink M, and Konstantinidis E. (2015) Uncertainty and Exploration in a Restless Bandit Problem. *Topics in Cognitive Science* 7(2): 351-367.

Tang L, Jiang Y, Li L, and Li T (2014) Ensemble Contextual Bandits for Personalized Recommendation. *Proceedings of the 8th ACM Conference on Recommender Systems* (ACM), 73-80.

Vozalis E, and Margaritis K G. (2003) Analysis of Recommender Systems Algorithms. *The 6th Hellenic European Conference on Computer Mathematics & Its Applications*: 732-745.

Wang Y-C, Kraut R, and Levine J M (2012) To Stay or Leave?: The Relationship of Emotional and Informational Support to Commitment in Online Health Support Groups. *Proceedings of the ACM 2012 Conference on Computer Supported Cooperative Work* (ACM), 833-842.

WHO. (2018) Obesity and Overweight, Accessed May 01, 2019. *https://www.who.int/news-room/fact-sheets/detail/obesity-and-overweight*.

Yan L. (2018) Good Intentions, Bad Outcomes: The Effects of Mismatches between Social Support and Health Outcomes in an Online Weight Loss Community. *Production and Operations Management* 27(1): 9-27.

Yan L, and Tan Y. (2014) Feeling Blue? Go Online: An Empirical Study of Social Support among Patients. *Information Systems Research* 25(4): 690-709.

Zeng C, Wang Q, Mokhtari S, and Li T. (2016) Online Context-Aware Recommendation with Time Varying Multi-Armed Bandit. *Proceedings of the 22nd ACM SIGKDD International Conference on Knowledge Discovery and Data Mining*: 2025-2034.

Zhang S, Yao L, Sun A, and Tay Y. (2019) Deep Learning Based Recommender System: A Survey and New Perspectives. *ACM Computing Surveys (CSUR)* 52(1): 1-38.